\documentclass[lettersize,journal]{IEEEtran}
\usepackage{amsmath,amsfonts}
\usepackage{algorithmic}
\usepackage{algorithm}
\usepackage{array}
\usepackage[caption=false,font=normalsize,labelfont=sf,textfont=sf]{subfig}
\usepackage{textcomp}
\usepackage{stfloats}
\usepackage{url}
\usepackage{verbatim}
\usepackage{graphicx}
\usepackage{cite}
\usepackage{caption}
\usepackage{amsmath}
\usepackage{hyperref}
\usepackage{amssymb}
\usepackage{mathtools}
\usepackage{amsthm}
\usepackage{microtype}
\usepackage{graphicx}
\usepackage{booktabs} 
\usepackage[capitalize,noabbrev]{cleveref}
\usepackage{pifont}
\newcommand{\gD}{\mathcal{D}}
\newcommand{\gC}{\mathcal{C}}
\def\cL{{\mathcal{L}}}

\usepackage{amsmath}  
\usepackage{amsfonts} 
\usepackage{amsmath}
\usepackage{multirow}
\begin{document}

\title{Few-Shot Vision-Language Action-Incremental Policy Learning}

\author{Mingchen Song, Xiang Deng*, Guoqiang Zhong,~\IEEEmembership{Member,~IEEE}, Qi Lv, Jia Wan, Yinchuan Li,~\IEEEmembership{Member,~IEEE}, Jianye Hao,~\IEEEmembership{Senior Member,~IEEE}, Weili Guan*,~\IEEEmembership{Member,~IEEE}
\thanks{*Corresponding author}
\thanks{Mingchen Song, Xiang Deng, Qi Lv, Jia Wan and Weili Guan are with the School of Computer Science and Technology, Harbin Institute of Technology (Shenzhen), China, 518055 (e-mail: \{mingchens0905, xdeng2023, aopolin.ii, jiawan1998, honeyguan\}@gmail.com).}

\thanks{Guoqiang Zhong is with the Department of Computer Science and Technology,
Ocean University of China (email: gqzhong@ouc.edu.cn).}
\thanks{Yinchuan Li and Jianye Hao are with the Huawei Noah’s Ark Lab, China (e-mail: \{liyinchuan, haojianye\}@huawei.com).}
}

\markboth{ }%
{Shell \MakeLowercase{\textit{et al.}}: A Sample Article Using IEEEtran.cls for IEEE Journals}


\maketitle

\begin{abstract}
Recently, Transformer-based robotic manipulation methods utilize multi-view spatial representations and language instructions to learn robot motion trajectories by leveraging numerous robot demonstrations. However, the collection of robot data is extremely challenging, and existing methods lack the capability for continuous learning on new tasks with only a few demonstrations. In this paper, we formulate these challenges as the \textbf{F}ew-\textbf{S}hot \textbf{A}ction-\textbf{I}ncremental \textbf{L}earning (\textbf{FSAIL}) task, and accordingly design a \textbf{T}ask-pr\textbf{O}mpt gra\textbf{P}h evolut\textbf{I}on poli\textbf{C}y (\textbf{TOPIC}) to address these issues. Specifically, to address the data scarcity issue in robotic imitation learning, TOPIC learns \textbf{T}ask-\textbf{S}pecific \textbf{P}rompts (\textbf{TSP}) through the deep interaction of multi-modal information within few-shot demonstrations, thereby effectively extracting the task-specific discriminative information. On the other hand, to enhance the capability for continual learning on new tasks and mitigate the issue of catastrophic forgetting, TOPIC adopts a \textbf{C}ontinuous \textbf{E}volution \textbf{S}trategy (\textbf{CES}). CES leverages the intrinsic relationships between tasks to construct a task relation graph, which effectively facilitates the adaptation of new tasks by reusing skills learned from previous tasks. TOPIC pioneers few-shot continual learning in the robotic manipulation task, and extensive experimental results demonstrate that TOPIC outperforms state-of-the-art baselines by over 26$\%$ in success rate, significantly enhancing the continual learning capabilities of existing Transformer-based policies\footnote{Our code will be available at https://github.com/codeshop715/FSAIL.}.

\end{abstract}


\section{Introduction}
\begin{figure}[tph!]
	\centering  
	\includegraphics[width=0.45\textwidth]{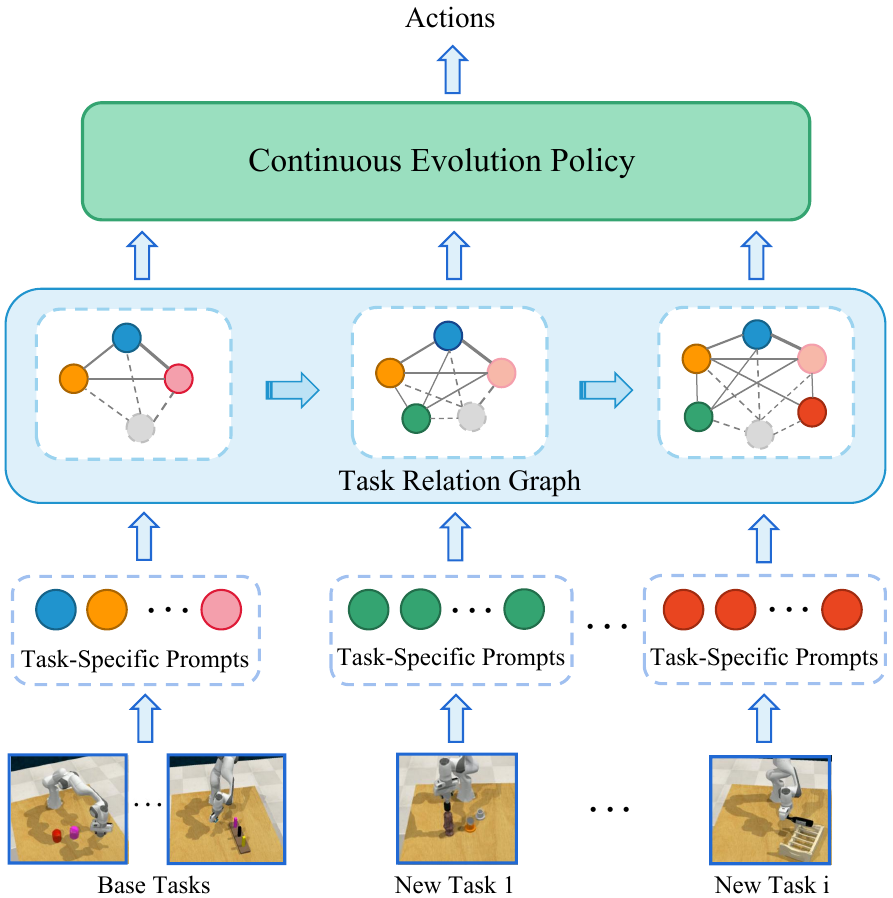}  
	\caption{Illustration of our proposed Task-prOmpt graPh evolutIon poliCy (TOPIC) for FSAIL. We learn task-specific prompts and construct a task relation graph with few-shot demonstrations. TOPIC has the ability to perform adaptive policy weights based on the intrinsic relationships between different tasks through a continuous evolution strategy.}  
	\label{fig:1}  
\end{figure}  

\IEEEPARstart{R}obotic manipulation represents a fundamental challenge in embodied AI~\cite{huang2023voxposer,huang2023instruct2act,li2024manipllm,ma2024hierarchical}, requiring systems to perform diverse physical interactions with objects in complex environments. These tasks~\cite{ xu2024dexterous,xu2023unidexgrasp,driess2023palm} demand precise control of end-effectors or joints while processing and integrating multi-modal information from both vision and language inputs to understand objectives and generate actions. 
In recent years, Transformer-based policies~\cite{jiang2022vima,li2023mastering,guhur2023instruction,goyal2023rvt,goyal2024rvt,zhang2024sam,chen2024sugar} have emerged as a powerful paradigm for robotic manipulation, showing exceptional versatility and dexterity across diverse scenarios. Specifically, these approaches leverage the Transformer architecture's attention mechanisms to effectively process multi-modal inputs and generate precise actions. By incorporating sophisticated spatial representations and language instructions, these models enable precise prediction of end-effector poses and complex manipulation capabilities across diverse environmental contexts.

In spite of these advancements that have significantly improved performance in robotic manipulation, there are still two critical issues. First, existing methods heavily rely on a massive quantity of high-quality robot demonstrations for imitation. However, it is very difficult and expensive to collect large amounts of robotic demonstration data in various environments. Second, the embodied tasks in the real world are diverse, and thus robotic systems need to continuously adapt to new tasks while avoiding forgetting previous tasks. Especially with just a few demonstrations on new tasks, current methods struggle to retain previously learned skills while learning new skills, leading to significant catastrophic forgetting. Consequently, it is crucial to enhance the continuous learning capabilities of existing methods on new tasks with few-shot demonstrations.

The above observations lead to two fundamental questions: \textbf{\ding{182}} \textit{How to fully exploit and utilize the information contained in few-shot demonstrations to learn an effective policy?} \textbf{\ding{183}} \textit{How to enhance the Transformer-based policy’s ability to continuously learn new tasks while preserving previously acquired skills?} To address these challenges, we propose the Few-Shot Action-Incremental Learning (FSAIL) task, and accordingly design a Task-prOmpt graPh evolutIon poliCy (TOPIC). TOPIC can be flexibly integrated with existing Transformer-based policies for robotic manipulation, enhancing their continuous learning capabilities with only a few demonstrations.

Specifically, to answer question \textbf{\ding{182}}, we introduce Task-Specific Prompts (TSP) that are suitable for embodied tasks. We predefine a set of learnable prompt vectors that can deeply interact with the multi-modal input data within few-shot demonstrations. By doing so, the task-specific prompts possess the capability to aggregate information from different modalities, thereby effectively extracting the task-specific information. Moreover, we further process this information through the task-specific prompt projection module to extract task-specific discriminative information to guide the prediction of actions. Our proposed task-specific prompts effectively alleviate the issue of data scarcity in embodied tasks.

In response to question \textbf{\ding{183}}, we propose a Continuous Evolution Strategy (CES) that can adapt to new tasks while preserving and leveraging the skills acquired from previous tasks, thereby fostering skill transfer from learned tasks to new tasks. For concreteness, we utilize the learned task-specific prompts to capture the intrinsic relationships between tasks and construct a task relation graph. As shown in Figure \ref{fig:1}, as the model continuously learns and adapts to new tasks, the task relation graph retains the intrinsic connections among various tasks. When adapting to a new task, we update the policy weights according to the task relation graph between different tasks, enabling existing Transformer-based policies to adapt to new tasks while mitigating the catastrophic forgetting of previously learned tasks.

Our contributions can be summarized as follows:
\begin{itemize}
\item Our work pioneers few-shot continual learning in robotic manipulation, introducing the \textbf{F}ew-\textbf{S}hot  \textbf{A}ction-\textbf{I}ncremental \textbf{L}earning (\textbf{FSAIL}) task and accordingly designing a \textbf{T}ask-pr\textbf{O}mpt gra\textbf{P}h evolut\textbf{I}on poli\textbf{C}y (\textbf{TOPIC}). TOPIC significantly enhances the continual learning capabilities of existing Transformer-based policies, enabling them to adapt to various new tasks while preserving the skills learned from previous tasks.
\item To address the issue of data scarcity,  we propose \textbf{T}ask-\textbf{S}pecific \textbf{P}rompts (\textbf{TSP}) that are suitable for embodied tasks. TSP have the capability to aggregate multi-modal information with few-shot demonstrations and extract discriminative information specific to the current task, guiding the prediction of actions.
\item To enhance the continual learning capability of existing methods, we introduce a \textbf{C}ontinuous \textbf{E}volution \textbf{S}trategy (\textbf{CES}). CES leverages the intrinsic relationships between tasks and construct a task relation graph, enabling TOPIC to adapt to new tasks by reusing skills learned from previous tasks.
\item Our work focuses on few-shot action continual learning for robotic manipulation tasks, effectively adapting to new tasks while mitigating catastrophic forgetting, setting a benchmark in the field. Extensive experiments in both simulation and real-world scenarios demonstrate the robustness and effectiveness of our proposed method.
\end{itemize}
\section{Related Work}
In this section, we provide a brief overview of recent advancements relevant to the proposed TOPIC, including few-shot imitation learning, class-incremental learning, and continual learning in robotics.
\subsection{Few-Shot Imitation Learning}
Few-shot imitation learning extends general few-shot learning paradigm~\cite{chen2023semantic,jeong2023winclip,zhu2023not,zhang2022tip,zhou2022conditional,yan2023smae,fu2023styleadv,huang2023rethinking,yang2025hyperbolic,wang2024cross,xi2025transductive,xu2024enhancing} to the domain of imitation learning, enabling robotic systems to efficiently learn and adapt to novel tasks from a limited number of demonstrations. This field encompasses three primary approaches: meta-learning, transfer learning, and 3D point cloud-based methods. Meta-learning approaches~\cite{duan2017one,finn2017one,james2018task,zhang2024one} incorporate established meta-learning techniques into embodied tasks, allowing models to acquire new skills from minimal demonstrations. A notable example is one-shot imitation learning~\cite{duan2017one}, which presents a meta-learning framework capable of learning from a single task demonstration and successfully generalizing to new instances of that task. Transfer learning approaches~\cite{nair2022r3m,bousmalis2023robocat,team2024octo,kim2024openvla} focus on training models with extensive datasets to develop comprehensive knowledge bases that can be effectively transferred to robotic manipulation tasks. For instance, R3M~\cite{nair2022r3m} develops robust visual representations through pretraining on large-scale human video datasets before fine-tuning with robot-specific data, creating a model particularly effective for robotic manipulation challenges. The third approach~\cite{goyal2023rvt,goyal2024rvt,shridhar2023perceiver,ze2023gnfactor} leverages 3D point cloud technology to maximize the utilization of spatial information from limited demonstrations. PERACT~\cite{shridhar2023perceiver} exemplifies this approach by employing the Perceiver Transformer architecture to encode language goals and RGB-D voxel patches, then generating discretized actions by identifying optimal voxel-based interventions. In contrast to these methods, we introduce Task-Specific Prompts (TSP), a prompt learning approach specifically designed for embodied tasks in the robotic manipulation domain. TSP effectively extract task-specific discriminative features through deep multi-modal information integration from just a few demonstrations, thereby effectively guiding subsequent action prediction processes.
\subsection{Class-Incremental Learning}
Class-Incremental Learning (CIL) enables models to progressively acquire knowledge of new classes while maintaining previously learned information~\cite{zhou2023revisiting,wang2022foster,tao2020topology,li2024mamba,d2023multimodal,tao2020few,zhang2021few,zhou2022forward,zhuang2023gkeal,yan2021dynamically,liu2024ntk,li2024relationship,ji2023memorizing}. CIL approaches primarily fall into three categories: regularization-based, replay-based and prompt-based methods. Regularization-based techniques preserve existing knowledge by strategically constraining parameter updates during training. Notable examples include EWC~\cite{kirkpatrick2017overcoming}, MAS~\cite{aljundi2018memory}, and SI~\cite{zenke2017continual}, which quantify parameter importance and implement targeted regularization to safeguard critical weights from significant modifications. Replay-based methods maintain a subset of previously encountered training examples throughout the learning process to facilitate retention of prior knowledge while adapting new classes~\cite{rebuffi2017icarl,iscen2020memory,xiang2019incremental,smith2024adaptive,zhou2024balanced}. By selectively preserving and revisiting representative samples from earlier learning stages, these approaches enable models to maintain a nuanced understanding of previous task distributions. Prompt-based methods~\cite{smith2023coda,wang2024hierarchical,wang2022dualprompt,wang2022s} learn a set of trainable prompt vectors to improve the model's continual learning performance. For example, S-Prompts~\cite{wang2022s} proposes an independent prompt learning paradigm that trains prompts separately for each domain using pre-trained transformers, avoiding the common requirement for exemplar storage. Differing from existing techniques, we introduce task-specific prompts and build a task relation graph using few-shot demonstrations. TOPIC has the ability to perform adaptive policy weights based on the intrinsic relationships between different tasks through a continuous evolution strategy, demonstrating notable capabilities in adapting to novel tasks through few-shot learning paradigms while effectively mitigating catastrophic forgetting.
\subsection{Continual Learning in Robotics}
Continual learning in robotics presents significant challenges~\cite{ayub2022few}. 
Recent work has explored continual learning in Vision-Language Navigation (VLN). VLNCL~\cite{li2024vision} propose a novel dual-loop scenario replay method for VLN agents that organizes and replays task memories to adapt to new environments while mitigating catastrophic forgetting. Similarly, CVLN~\cite{jeong2024continual} introduce the continual VLN paradigm with two novel rehearsal-based methods: perplexity replay for prioritizing challenging episodes and episodic self-replay for maintaining learned behaviors via action logits replay.
For manipulation tasks, several approaches have been developed on the LIBERO benchmark~\cite{liu2023libero}, which focuses on knowledge transfer for lifelong robot learning. LOTUS~\cite{wan2024lotus} constructs an ever-growing skill library through continual skill discovery using open-vocabulary vision models to extract recurring patterns from unsegmented demonstrations. M2Distill~\cite{roy2024m2distill} preserves consistent latent space across vision, language, and action distributions throughout the learning process. TAIL~\cite{liu2023tail} explores parameter-efficient fine-tuning techniques to adapt large pretrained models for new control tasks in continual learning settings.
In contrast to these works, we aim to address the more challenging Few-Shot Action-Incremental Learning (FSAIL) task where the new task has only a few demonstrations, which is significantly different from other works. This setting not only requires models to possess the ability to learn skills from only one or five demonstrations, but also demands that models continuously learn new tasks while avoiding catastrophic forgetting of previously learned skills. Our work pioneers few-shot continual learning in robotic manipulation, significantly enhancing the continuous learning capabilities of existing Transformer-based policies.
\begin{figure*}[t] 
	\centering 
	\includegraphics[width=0.97\textwidth]{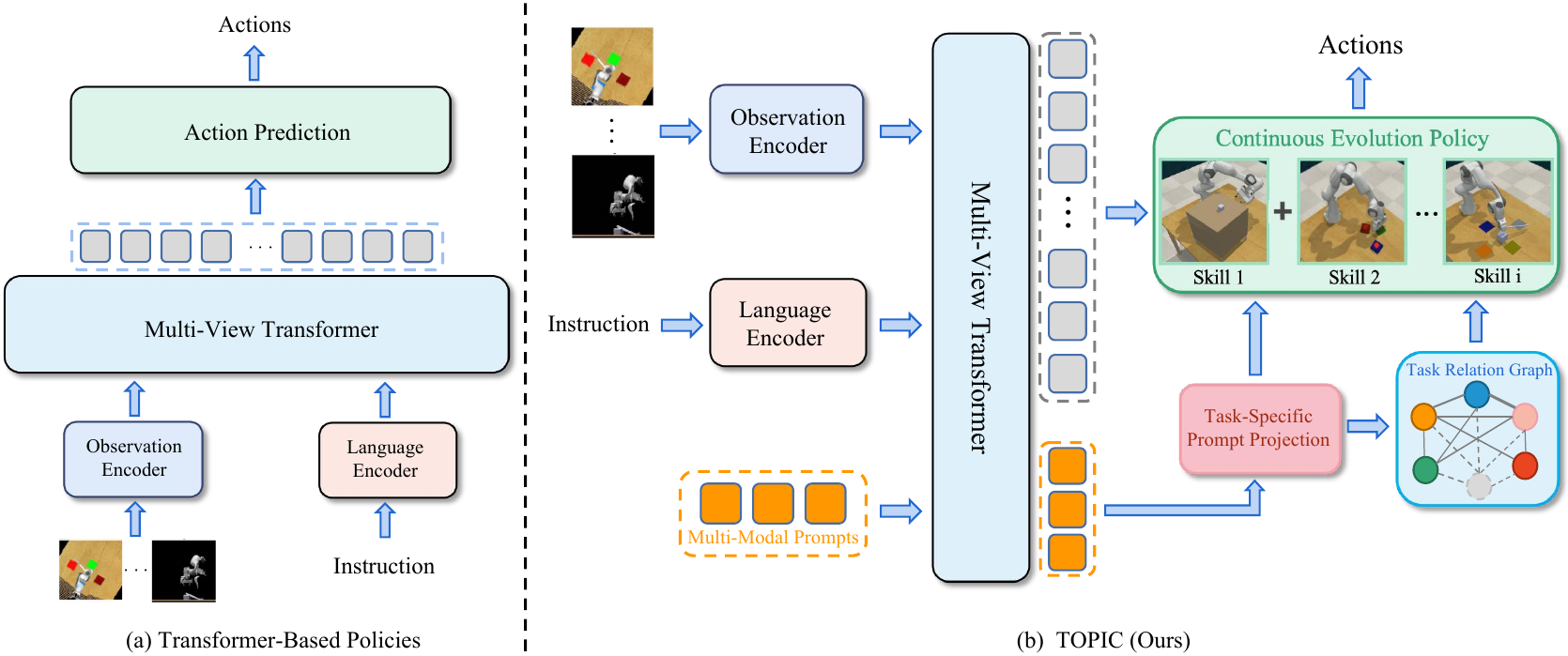} 
    \caption{Comparisons of different Transformer-based policies and our proposed TOPIC. (a): Transformer-based policies include a series of methods such as RVT, RVT2, SAM-E, and others. (b): Our proposed TOPIC, which can be flexibly integrated with other Transformer-based policies to enhance their continual learning capability with few-shot demonstrations.} 
	\label{fig:2} 
\end{figure*}
\section{Method}
In this section, we first introduce the task definition of Few-Shot Action-Incremental Learning (FSAIL) in Section ~\ref{sec:FSAIL}, and then outline the overall architecture of our proposed Task-prOmpt graPh evolutIon poliCy (TOPIC) in Section ~\ref{sec:Overview}. The proposed Task-Specific Prompts (TSP), Continuous Evolution Strategy (CES) and training procedure are presented in Section ~\ref{sec:TSP}, Section~\ref{sec:CES} and Section~\ref{sec:Training}, respectively.
\subsection{Few-Shot Action-Incremental Learning}
\label{sec:FSAIL}
FSAIL aims to empower embodied models with the capability to perform continuous learning with just a few demonstrations. Specifically, FSAIL trains a model incrementally in multiple sessions $\{\gD^{(0)}, \gD^{(1)}, \dots, \gD^{(T)}\}$, where $\gD^{(t)}=\{(x_i,y_i)\}_{i=1}^{|\gD^{(t)}|}$ represents the training set for session $t$. $\gD^{(0)}$ is the base session, and $T$ is the number of incremental sessions. The base session $\gD^{(0)}$ contains extensive training data for each task $c\in\gC^{(0)}$. In each incremental session $\gD^{(t)}$, $t>0$, there are only a few demonstrations data, $|\gD^{(t)}|=p \times q$, where $p$ is the number of tasks and $q$ is the number of demonstrations per unseen task, We define FSAIL tasks as $q$-shot task based on the number of demonstrations available for each task in the incremental session. The training sets from previous sessions are not accessible, which requires the model to generalize to new tasks without forgetting previously learned skills. Evaluation in session $t$ involves test data from all tasks encountered up to that session, \emph{i.e.,} the action space of $\cup_{i=0}^{t}\gC^{(i)}$.
\subsection{Overview of the TOPIC for FSAIL}
\label{sec:Overview}
In order to adapt to new tasks with a few demonstrations while avoiding catastrophic forgetting, we propose the Task-prOmpt graPh evolutIon poliCy (TOPIC). TOPIC learns task-specific prompts (TSP) through the deep interaction of multi-modal information and adopts a Continuous Evolution Strategy (CES) that enables adaptation to new tasks by reusing skills from previous tasks. Notably, TOPIC can be flexibly integrated into existing Transformer-based policies and significantly enhances their continual learning abilities. Figure \ref{fig:2} provides a comparison of our proposed TOPIC with other Transformer-based policies. 
\begin{figure}[t]
	\centering  
	\includegraphics[width=0.417\textwidth]{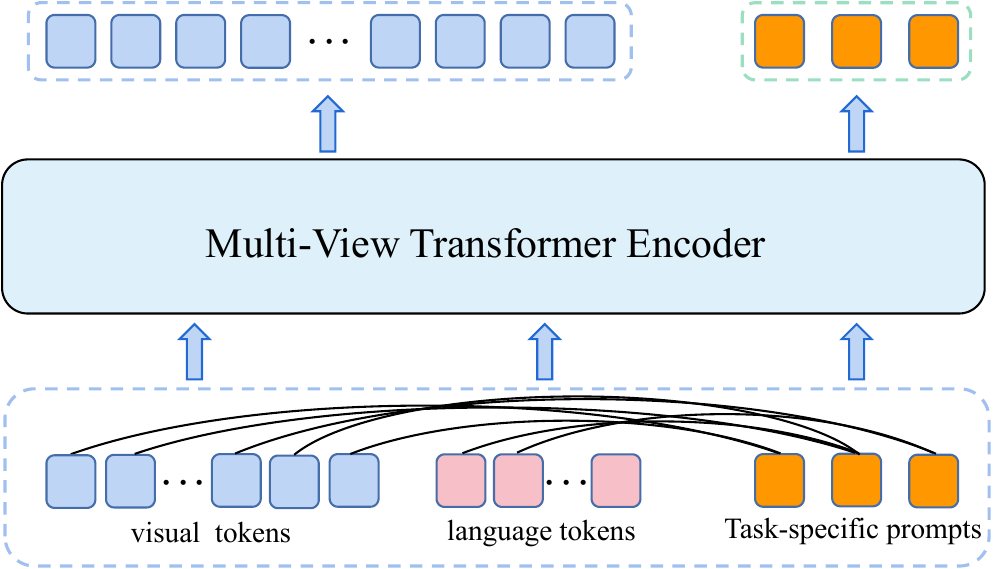}  
	\caption{The structure of our proposed Task-Specific Prompts (TSP) involves a set of predefined learnable prompt vectors, which interact deeply with information from other modalities through the Multi-View Transformer Encoder. TSP extracts task-specific discriminative information within a few demonstrations.}  
	\label{fig:3}  
\end{figure} 
\subsection{Task-Specific Prompts}
\label{sec:TSP}
To mitigate data scarcity challenges in embodied tasks, we propose Task-Specific Prompts (TSP), a novel approach tailored to few-shot continual learning in robotic manipulation. We predefine a set of learnable task-specific prompts $\textbf{P} \in \mathbb{R}^{n \times \textit{C}}$ that are randomly initialized, which are subsequently processed through a Multi-View Transformer Encoder to enable deep cross-modal interactions. TSP extracts task-specific discriminative information within a few demonstrations. In particular, the language tokens $\textbf{T} \in \mathbb{R}^{m \times \textit{C}}$ processed by the text encoder and the visual tokens $\textbf{O} \in \mathbb{R}^{k \times \textit{C}}$ processed by the visual encoder are represented as follows:
\begin{equation}
\begin{cases}
\textbf{P} = \{ \textit{\textbf{p}}_1, \textit{\textbf{p}}_2, \dots, \textit{\textbf{p}}_n \},\\
\textbf{T} = \{ \textit{\textbf{t}}_1, \textit{\textbf{t}}_2, \dots, \textit{\textbf{t}}_m \},\\
\textbf{O} = \{ \textit{\textbf{o}}_1, \textit{\textbf{o}}_2, \dots, \textit{\textbf{o}}_k \},
\end{cases}
\end{equation}
where $\textit{n}$ denotes the number of task-specific prompts, $\textit{m}$ represents the number of language tokens, and $\textit{k}$ signifies the number of visual tokens, all of which share the same dimensionality $\textit{C}$. We concatenate task-specific prompts with the text and visual tokens to obtain the input $\textbf{X} \in \mathbb{R}^{(n + m + k) \times \textit{C}}$ for the Multi-View Transformer~\cite{guhur2023instruction}:
\begin{equation}
 \textbf{X} = \text{concatenate}(\textbf{P}, \textbf{T}, \textbf{O}).
\end{equation}
Then, the token sequence $\textbf{X}$, containing information from different modalities, is fed into the Multi-View Transformer layers that can simultaneously receive visual and language modality tokens. As shown in Figure \ref{fig:3}, this process facilitates deep interaction between task-specific prompts, visual, and language tokens. By doing so, TSP aggregate task-specific information from a few demonstrations across different modalities relevant to the current task. Concretely, the self-attention~\cite{vaswani2017attention} calculation can be represented as follows:
\begin{equation}
 \textbf{Q}, \textbf{K},\textbf{V}= \textbf{W}_{Q}\textbf{X},\textbf{W}_{K}\textbf{X},\textbf{W}_{V}\textbf{X},
\end{equation}
\begin{equation}
      \text{Attention}(\textbf{Q}, \textbf{K}, \textbf{V})= \text{Softmax}(\frac{\textbf{Q}(\textbf{K})^{T}}{\sqrt{d_k}})\textbf{V}.
\end{equation}

After processing through the Multi-View Transformer Encoder (MVTE), we obtain feature representations $\hat{\textbf{X}}$ and task-specific prompts $\hat{\textbf{P}}$. The feature extraction process can be formalized as follows:
\begin{equation}
  \hat{\textbf{X}}, \hat{\textbf{P}} =  \text{MVTE}(\textbf{X}),
\end{equation}
where $\textbf{X}$ represents the input features. These prompts $\hat{\textbf{P}}$ are projected to match the dimensionality of $\hat{\textbf{X}}$ via a task-specific prompt projection module denoted as $h$. The projection can be implemented in three ways: identity mapping, linear transformation, or MLP projection. The task-specific output features $\textbf{X}_{out}$ are then generated by combining the extracted features with the projected prompt through a broadcast addition:
\begin{equation}
 \textbf{X}_{out} = \hat{\textbf{X}} + h(\hat{\textbf{P}}).
\end{equation}
This approach allows the model to incorporate task-specific information directly into the action space, enhancing performance on targeted tasks.
\subsection{Continuous Evolution Strategy}
\label{sec:CES}
Intuitively, skills are systematically reused across different tasks~\cite{wang2024sparse}. For example, ``pick and place" is a common skill frequently utilized in robotic tasks. Moreover, skills required for new tasks can be integrated into the existing learned skills, enabling their flexible application in subsequent tasks. Inspired by this, we propose a Continuous Evolution Strategy (CES), which can learn new tasks while utilizing the skills acquired from previous tasks to update the policy weights $\textbf{W}$, thereby facilitating the transfer of skills learned from previous tasks to new tasks. Specifically, we leverage the task-specific prompts $\hat{\textbf{P}}$ extracted in the previous section to represent task-specific discriminative information. Each task has its own dedicated Task-Specific Prompt $\hat{\textbf{P}}_i$ and its corresponding policy weight $\textbf{W}_i$ :
\begin{equation}
(\hat{\textbf{P}}_i, \textbf{W}_i), \quad i=1,2,\dots,n.
\label{eq:7}
\end{equation}
Then, we construct a task relation graph to model the intrinsic relationships between tasks, where each task-specific prompts $\hat{\textbf{P}}_i$ can be regarded as a graph node. Notably, the graph structure possesses several desirable properties that make it an appropriate tool for representing the intrinsic relationships between tasks. 
First, as the graph nodes are continuously updated, the number of reusable skills increases accordingly, which benefits the model's adaptation to new tasks.
Secondly, the graph structure allows a trained model to be extended to any number of tasks, meaning that the updating of policies at any sessions can share the same learned task relation graph.
To illustrate how the intrinsic relationships between tasks are utilized for weight updating, we take the updating of $\textit{node}_j$ in the graph as an example. We first compute a relation coefficient $\textit{s}_{ij}$ between $\textit{node}_j$ and all nodes in the graph, such as $\textit{node}_i$ and $\textit{node}_j$:
\begin{equation}
s_{ij} = d( \hat{\textbf{P}}_i,\hat{\textbf{P}}_j ),\\
\end{equation}
where $d( \hat{\textbf{P}}_i,\hat{\textbf{P}}_j )$ is the cosine distance between the task-specific prompts $\hat{\textbf{P}}_i$ for task $\textit{i}$ and  the task-specific prompts $\hat{\textbf{P}}_j$ for task $\textit{j}$. We represent the intrinsic relationships information between tasks in the task relation graph based on $s_{ij}$, and aggregate it with the current task’s policy weights to obtain the updated weights $\textbf{W}_j$:
\begin{equation}
\hat{\textbf{W}}_j = \frac{1}{j-1}\sum_{i = 1}^{j-1}s_{ij}\textbf{W}_i + \textbf{W}_j.\\
\end{equation}
In other words, the policy weights $\textbf{W}_j$ for the new task $j$ are derived from a combination of the policy weights of all previous tasks. Meanwhile, in order to retain the general skill shared among tasks, we introduce the common skill weight $\textbf{W}_{base}$ from the base session. Therefore, the final weight update for the policy $\hat{\textbf{W}}_j$ can be formulated as:
\begin{equation}
\hat{\textbf{W}}_j = \lambda_1 \left(\frac{1}{j-1}\sum_{i = 1}^{j-1}s_{ij}\textbf{W}_i + \textbf{W}_j\right) + \lambda_2\textbf{W}_{base},\\
\end{equation}
where we regulate the weight of task-specific skills through coefficient $\lambda_1$, and adjust the weight of general skills learned in the base task through coefficient $\lambda_2$. Therefore, in each incoming session, we leverage the intrinsic relationships between different tasks to update the policy weights in the current session. Subsequently, we utilize the updated policy to make action $a$ predictions across all previous tasks and  the current task $j$:
\begin{equation}
    {a}_{i=1}^j = \hat{\textbf{W}}_j(\textbf{X}_{out})_{i=1}^j.
\end{equation}
\subsection{Training Procedure}
\label{sec:Training}
Our training process is divided into three stages. In the first stage, we employ a multi-task training approach to train the model backbone $\textbf{R}$ and policy weight $\textbf{W}_{base}$ in the base session, aiming to acquire the general skill. Subsequently, we train each base class task separately to obtain task-specific prompts and policy weights in Equation~\ref{eq:7}. In the incremental learning stage, we use few-shot training data to train in new sessions. During the latter two stages, the text and visual encoders remain frozen. We adopt the fundamental paradigm of imitation learning~\cite{RoboFlamingo} to address language-conditioned manipulation tasks. Specifically, imitation learning enables the model to mimic a set of expert demonstrations denoted as $\gD:=\{(\tau, l)_{i}\}_{i=0}^{|\gD|}$, where $\tau:=(o_0,a_0,\dots,o_{T-1},a_{T-1},o_T)$ represents the expert trajectory, and $l$ denotes the language instruction. A common imitation learning objective for the model weight $\pi_{\theta}$ is to maximize the likelihood of actions conditioned on the language and current state:
\begin{equation}
\label{eq:imitation}
\cL(\theta):=-\mathbb{E}_{(\tau,l)\sim \gD}\left[\sum_{t=0}^{T-1}\log\pi_{\theta}(a_t|o_t,l)\right].
\end{equation}
\section{Experiments}
In this section, we first introduce the experimental settings, followed by a comparison of the results with state-of-the-art methods. Finally, we conduct ablation studies and detailed analysis to validate the effectiveness of our method.
\subsection{Experiment Setup}
\subsubsection{Baselines}  
Since our proposed method significantly enhances the continual learning capability of the Transformer-based policies, we have chosen three general Transformer-based policies, including: (i): RVT~\cite{goyal2023rvt} is the state-of-the-art multi-view architecture for 3D manipulation, which re-renders visual observations into orthographic projections of cube views and predicts the next move based on these projections. (ii): SAM-E~\cite{zhang2024sam} improves RVT by integrating the SAM encoder and predicting action sequences. (iii): RVT-2~\cite{goyal2024rvt} is a single model capable of performing multiple 3D manipulation tasks, including those that require millimeter-level precision.
\subsubsection{Continuous Learning Methods} 
For a fair comparison, we apply the replay-based, regularization-based and prompt-based continual learning methods to the existing Transformer-based policies. First, for the replay-based~\cite{smith2024adaptive} method, we replay demonstrations of the incremental tasks during the incremental learning phase. Second, for the regularization-based method~\cite{kirkpatrick2017overcoming}, we determine the importance of parameters by calculating the magnitude of the gradients, selectively updating the parameters accordingly. Third, we integrate the state-of-the-art prompt-based continual learning method S-prompts~\cite{wang2022s} with existing Transformer-based policies, which leverages both visual and language prompts.
\begin{figure*}[t] 
	\centering 
	\includegraphics[width=1.0\textwidth]{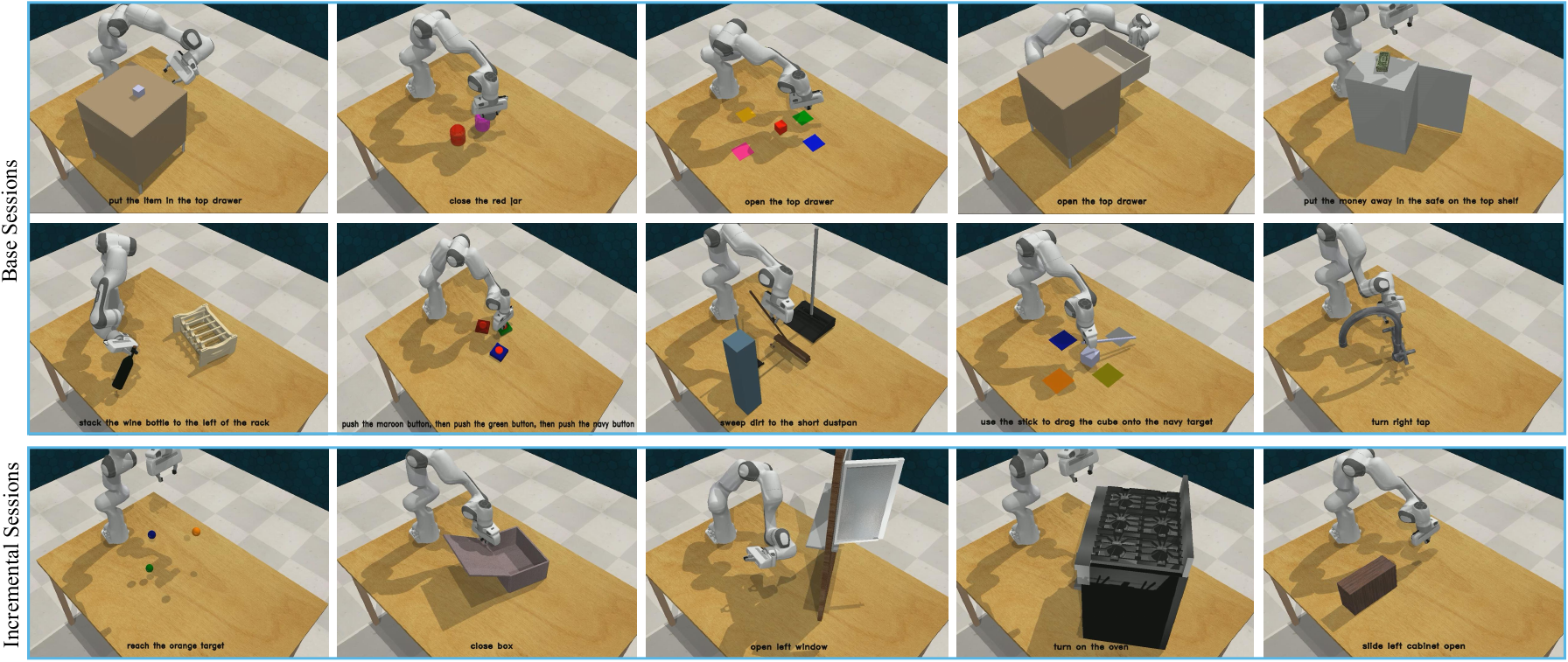} 
	\caption{FSAIL Tasks in RLBench. We design 10 tasks in the base session and 5 tasks in the incremental session to validate the model's continual learning capability with novel objects and actions.}
	\label{fig:8} 
\end{figure*}
\begin{table*}[t]
	\renewcommand\arraystretch{1.1}
	\begin{center}
		\centering
		\caption{FSAIL performance across sessions on \textbf{1-shot} tasks compared with other methods. ``Average Acc." is the average accuracy across all sessions. ``Final Improv." denotes the average accuracy improvement compared to baseline. Mean and std of 5 evaluations are reported.}
		\resizebox{1\textwidth}{!}{
  
			\begin{tabular}{lcccccccl}
				\toprule
				\multicolumn{1}{l}{\multirow{2}{*}{\bf Methods}}&\multicolumn{6}{c}{\bf Accuracy in each session (\%) $\uparrow$}&\bf Average&\bf Final \\ 
				\cmidrule{2-7}
			      &\bf 0&\bf 1&\bf 2&\bf 3&\bf 4&\bf 5&\bf Acc.&\bf Improv.  \\ 
				\midrule
                     \multicolumn{9}{l}{\emph{Transformer-based policies}} \\
				RVT~\cite{goyal2023rvt}  &80.2 ± 0.5&17.1 ± 0.7&14.3 ± 0.5&11.2 ± 0.4&10.6 ± 0.5&9.5 ± 0.4&23.8 ± 0.5& baseline\\
                    SAM-E~\cite{zhang2024sam} & 91.6 ± 0.3&26.3 ± 0.6&24.4 ± 0.9&18.2 ± 0.3&11.4 ± 0.6&10.9 ± 1.0&30.5 ± 0.6& baseline\\
				RVT-2~\cite{goyal2024rvt} &91.6 ± 1.2&32.1 ± 0.8&28.2 ± 1.1&24.3 ± 0.9&17.1 ± 0.6&11.7 ± 0.4&34.2 ± 0.8& baseline\\
                    \midrule
                    \multicolumn{9}{l}{\emph{Transformer-based policies + Continuous Learning methods}}\\
                    RVT~\cite{goyal2023rvt} + Replay~\cite{smith2024adaptive} &80.2 ± 0.5&17.1 ± 0.7&16.6 ± 0.5&16.4 ± 0.6&17.3 ± 0.5&16.0 ± 0.5&27.3 ± 0.6& + 3.5\\
                    SAM-E~\cite{zhang2024sam} + Replay~\cite{smith2024adaptive}& 91.6 ± 0.3&26.3 ± 0.6&26.1 ± 0.1&25.5 ± 0.6&24.0 ± 1.0&23.8 ± 2.1&36.2 ± 0.8& + 5.7\\
			    RVT-2~\cite{goyal2024rvt}  + Replay~\cite{smith2024adaptive} &91.6 ± 1.2&32.1 ± 0.8&31.8 ± 1.7&31.4 ± 0.7&30.2 ± 0.8&30.6 ± 0.7&41.3 ± 1.0& + 7.1\\
			   RVT~\cite{goyal2023rvt} + Regularization~\cite{kirkpatrick2017overcoming} &80.2 ± 0.5&29.3 ± 0.9&24.4 ± 0.7&22.4 ± 0.7&20.5 ± 0.6&18.9 ± 1.5&32.6 ± 0.8& + 8.8\\
                    SAM-E~\cite{zhang2024sam} + Regularization~\cite{kirkpatrick2017overcoming}  & 91.6 ± 0.3&36.2 ± 0.2&34.3 ± 0.3&32.7 ± 1.2&29.4 ± 0.9&26.5 ± 0.1&41.8 ± 0.5& + 11.3\\
			    RVT-2~\cite{goyal2024rvt}  + Regularization~\cite{kirkpatrick2017overcoming} &91.6 ± 1.2&40.4 ± 1.6&37.7 ± 0.9&34.8 ± 0.3&34.7 ± 0.4&28.7 ± 0.3&44.7 ± 0.8& + 10.5\\
                RVT~\cite{goyal2023rvt} + S-prompts~\cite{wang2022s} &80.7 ± 0.3&31.6 ± 1.0&29.3 ± 0.9&25.8 ± 0.8&22.6 ± 1.2&19.2 ± 1.6 &34.9 ± 1.0 & + 11.1\\
                    SAM-E~\cite{zhang2024sam} + S-prompts~\cite{wang2022s}& 91.7 ± 0.4&41.1 ± 1.1&36.4 ± 1.2&29.7 ± 1.5&25.8 ± 1.1&22.8 ± 1.2& 41.3 ± 1.1 & + 10.8\\
			    RVT-2~\cite{goyal2024rvt}  + S-prompts~\cite{wang2022s} &91.8 ± 1.1&44.3 ± 0.9&39.0 ± 1.3&32.1 ± 1.0&28.1 ± 0.3&25.6 ± 1.2& 43.5 ± 1.0& + 9.3\\
			    
                    \midrule
				\multicolumn{9}{l}{\emph{\textbf{Ours}: Transformer-based policies + \textbf{TOPIC}} }\\
\multicolumn{1}{l}{RVT~\cite{goyal2023rvt} + \textbf{TOPIC}}&\textbf{81.3 ± 0.2}&\textbf{69.9 ± 0.7}&\textbf{51.9 ± 0.9}&\textbf{36.4 ± 0.8}&\textbf{29.5 ± 0.7}&\textbf{25.3 ± 1.2}&\textbf{49.1 ± 0.8}& \textbf{+ 25.3}
\\
\multicolumn{1}{l}{SAM-E~\cite{zhang2024sam} + \textbf{TOPIC} } &\textbf{91.9 ± 0.9}&\textbf{73.9 ± 1.4}&\textbf{64.8 ± 0.9}&\textbf{49.3 ± 0.3}&\textbf{39.5 ± 1.2}&\textbf{32.7 ± 0.8}&\textbf{58.7 ± 0.9}& \textbf{+ 28.2}
\\
\multicolumn{1}{l}{RVT-2~\cite{goyal2024rvt}  + \textbf{TOPIC} } &\textbf{91.8 ± 1.1}&\textbf{76.5 ± 0.9}&\textbf{61.3 ± 0.4}&\textbf{52.5 ± 0.3}&\textbf{43.8 ± 0.2}&\textbf{37.8 ± 0.5}&\textbf{60.6 ± 0.6}& \textbf{+ 26.4}
\\
\bottomrule
\end{tabular}
		}
		\label{table:1}
	\end{center}
\end{table*}
\subsubsection{Simulation Environment}
We conduct the experiments on a standard manipulation benchmark developed in RLBench~\cite{james2020rlbench}. A Franka Panda robot with a parallel gripper is controlled to complete the tasks. Visual observations are captured through four RGB-D cameras with a resolution of 128 × 128 (left shoulder, right shoulder, front, and wrist), and the target gripper pose is achieved leveraging a sample-based motion planner. For each task, there are 25 unseen demonstrations provided for testing.
\subsubsection{FSAIL Dataset Construction and Task Definition}
Considering the real-world application scenario, we construct a new dataset for FSAIL and corresponding evaluation metrics to explore the continual learning capability of Transformer-based policies. To fairly compare the continual learning capabilities of different methods, as shown in Figure \ref{fig:8}, we utilize the original 10 tasks from RLBench as the base session (with 100 demonstrations per task) and five newly generated incremental tasks. It is worth noting that the incremental session includes objects and actions different from those in the base session. We separately set up 1-shot and 5-shot FSAIL tasks, where $k$-shot represents the number of demonstrations available for new tasks during the incremental phase.
\begin{table*}[t]
	\renewcommand\arraystretch{1.1}
	\begin{center}
		\centering
		\caption{FSAIL performance across sessions on \textbf{5-shot} tasks compared with other methods. ``Average Acc." is the average accuracy across all sessions. ``Final Improv." denotes the average accuracy improvement compared to baseline. Mean and std of 5 evaluations are reported.}
		\resizebox{1\textwidth}{!}{
			\begin{tabular}{lcccccccl}
				\toprule
				\multicolumn{1}{l}{\multirow{2}{*}{\bf Methods}}&\multicolumn{6}{c}{\bf Accuracy in each session (\%) $\uparrow$}&\bf Average&\bf Final \\ 
				\cmidrule{2-7}
			      &\bf 0&\bf 1&\bf 2&\bf 3&\bf 4&\bf 5&\bf Acc.&\bf Improv.  \\ 
				\midrule
                     \multicolumn{9}{l}{\emph{Transformer-based policies}} \\
				RVT~\cite{goyal2023rvt}&80.2 ± 0.5&16.5 ± 1.2&15.4 ± 0.4&13.1 ± 1.4&11.3 ± 0.1&9.5 ± 0.8&24.3 ± 0.7& baseline\\
                    SAM-E~\cite{zhang2024sam} &91.6 ± 0.3&25.3 ± 0.7&21.0 ± 0.5&14.1 ± 0.4&11.7 ± 0.2&10.1 ± 0.7 &29.0 ± 0.5 & baseline\\
				RVT-2~\cite{goyal2024rvt} &91.6 ± 1.2&30.4 ± 0.3&27.1 ± 0.2&24.4 ± 2.1&21.5 ± 0.2&16.3 ± 0.2&35.2 ± 0.7& baseline\\
                    \midrule
                    \multicolumn{9}{l}{\emph{Transformer-based policies + Continuous Learning methods }}\\
                    RVT~\cite{goyal2023rvt} + Replay~\cite{smith2024adaptive} &80.2 ± 0.5&16.5 ± 1.2&15.8 ± 0.4&15.3 ± 1.4&14.4 ± 0.4&14.1 ± 0.1&26.1 ± 0.7& + 1.8\\
                    SAM-E~\cite{zhang2024sam} + Replay~\cite{smith2024adaptive} &91.6 ± 0.3&25.3 ± 0.7&24.2 ± 0.9&24.0 ± 0.6&21.7 ± 0.2&20.2 ± 1.1&34.5 ± 0.6& + 5.5\\
			    RVT-2~\cite{goyal2024rvt} + Replay~\cite{smith2024adaptive} &91.6 ± 1.2&30.4 ± 0.3&30.8 ± 0.2&28.2 ± 0.9&26.1 ± 0.4&25.8 ± 1.9&38.8 ± 0.8&+ 3.6\\
			    RVT~\cite{goyal2023rvt} + Regularization~\cite{kirkpatrick2017overcoming} &80.2 ± 0.5&28.4 ± 1.5&25.5 ± 0.8&24.9 ± 1.0&23.8 ± 0.3&23.1 ± 1.4&34.3 ± 0.9& + 10.0\\
                    SAM-E~\cite{zhang2024sam} + Regularization~\cite{kirkpatrick2017overcoming} & 91.6 ± 0.3&34.2 ± 1.2&33.3 ± 0.2&30.1 ± 0.5&29.4 ± 0.2&28.1 ± 0.2&41.1 ± 0.4& + 12.1\\
			    RVT-2~\cite{goyal2024rvt} + Regularization~\cite{kirkpatrick2017overcoming} &91.6 ± 1.2&40.1 ± 0.8&37.4 ± 0.7&36.0 ± 0.5&34.1 ± 0.8&33.5 ± 0.9&45.5 ± 0.8& + 8.9\\
                 RVT~\cite{goyal2023rvt} + S-prompts~\cite{wang2022s} &80.6 ± 0.7&30.9 ± 0.8&28.7 ± 0.7&24.1 ± 1.1&21.2 ± 1.6&20.7 ± 1.0 &34.4 ± 1.0 & + 10.1 \\
                    SAM-E~\cite{zhang2024sam} + S-prompts~\cite{wang2022s}& 91.7 ± 0.8&41.0 ± 1.2&35.2 ± 0.8&28.3 ± 1.0&26.4 ± 0.6&24.5 ± 0.4& 41.2 ± 0.8& + 12.2\\
			    RVT-2~\cite{goyal2024rvt}  + S-prompts~\cite{wang2022s} &91.8 ± 0.9&43.1 ± 0.8&37.2 ± 1.1&30.9 ± 0.5&29.3 ± 0.7&27.2 ± 1.8& 43.3 ± 1.0& + 8.1\\
			    
                    \midrule
				\multicolumn{9}{l}{\emph{\textbf{Ours}: Transformer-based policies + \textbf{TOPIC}} }\\
\multicolumn{1}{l}{RVT~\cite{goyal2023rvt} + \textbf{TOPIC}}&\textbf{81.3 ± 0.2}&\textbf{60.3 ± 1.8}&\textbf{50.4 ± 0.0}&\textbf{37.6 ± 0.4}&\textbf{33.2 ± 1.6}&\textbf{23.5 ± 0.3}&\textbf{47.7 ± 0.7}& \textbf{+ 23.4}
\\
\multicolumn{1}{l}{SAM-E~\cite{zhang2024sam} + \textbf{TOPIC} } &\textbf{91.9 ± 0.9}&\textbf{67.6 ± 1.7}&\textbf{64.1 ± 1.1}&\textbf{55.3 ± 0.9}&\textbf{43.1 ± 0.4}&\textbf{28.4 ± 0.3}&\textbf{58.4 ± 0.9}& \textbf{+ 29.4}\\
\multicolumn{1}{l}{RVT-2~\cite{goyal2024rvt} + \textbf{TOPIC} } &\textbf{91.8 ± 1.1}&\textbf{70.4 ± 0.3}&\textbf{66.4 ± 0.5}&\textbf{58.2 ± 1.4}&\textbf{44.1 ± 0.8}&\textbf{30.3 ± 0.4}&\textbf{60.2 ± 0.8}& \textbf{+ 25.0}
\\
				\bottomrule
			\end{tabular}
		}
		\label{table:2}
	\end{center}
\end{table*}
\begin{table}[t]
    \renewcommand{\arraystretch}{1.3}
    \setlength{\tabcolsep}{2.2pt}
    \centering
     \caption{Ablation study of our proposed TSP and CES.  We analyze the effectiveness of each component in the 1-shot task.  }
    \scalebox{0.98}{
    \begin{tabular}{lcccccccc}
        \toprule
        \multicolumn{1}{l}{\multirow{2}{*}{\bf Methods}} & \multicolumn{6}{c}{\bf Accuracy in each session (\%) $\uparrow$} &\bf Average &\bf Final\\ 
        \cmidrule(lr){2-7}
        & \bf 0 & \bf 1 & \bf 2 & \bf 3 & \bf 4 &\bf 5 & \bf Acc. &\bf Improv.\\
        \midrule
        \multicolumn{8}{l}{\emph{Transformer-based policies}} \\
        RVT &80.2 &17.1&14.3 &11.2 &10.6&9.5 &23.8  &baseline\\
         SAM-E & 91.6 &26.3 &24.4 &18.2 &11.4&10.9&30.5 &baseline\\
	RVT-2 &91.6 &32.1&28.2&24.3 &17.1&11.7 &34.2 &baseline\\
        \midrule
        \multicolumn{8}{l}{\emph{\textbf{Ours: Transformer-based policies + TSP}}} \\
        RVT + TSP      & 81.3 & 48.4  & 35.3  & 26.4 & 17.3 &14.5 &  37.2 &\textbf{+ 13.4} \\
         SAM-E  + TSP   & 91.9 & 56.6 & 46.6 & 37.4 & 24.0 & 16.8 & 45.6  &\textbf{+ 15.1}\\
       RVT-2 + TSP  & 91.8 &62.3 & 49.1 & 40.7 & 30.4 & 23.2 & 49.6 & \textbf{+ 15.4}\\
        \midrule
        \multicolumn{8}{l}{\emph{\textbf{Ours: Transformer-based policies + TSP \& CES} }} \\
        RVT + TSP \& CES & \textbf{81.3} & \textbf{69.9} & \textbf{51.9} & \textbf{36.4} & \textbf{29.5} & \textbf{25.3} & \textbf{49.1} & \textbf{+ 25.3} \\

         SAM-E + TSP \& CES & \textbf{91.9} & \textbf{73.9} & \textbf{64.8} & \textbf{49.3} & \textbf{39.5} & \textbf{32.7} & \textbf{58.7} & \textbf{+ 28.2} \\

        RVT-2  + TSP \& CES & \textbf{91.8} & \textbf{76.5} & \textbf{61.3} & \textbf{52.5} & \textbf{43.8} & \textbf{37.8} & \textbf{60.6} & \textbf{+ 26.4} \\
        \bottomrule
    \end{tabular}
                }
    \label{table:3}
\end{table}
\subsection{Comparison with the state-of-the-art methods}
We first train the model on 10 base tasks as our base session (session 0). Subsequently, we independently learn five unseen tasks in the incremental sessions and evaluate the model’s accuracy, which includes 1-shot and 5-shot FSAIL tasks. We present comparisons between our proposed method and other methods in Table~\ref{table:1} and Table~\ref{table:2}. Our method surpasses the current state-of-the-art methods both on 1-shot and 5-shot tasks. 
Specifically, compared to the RVT, SAM-E, and RVT-2 baselines, our proposed method outperforms existing Transformer-based policies substantially. As shown in Table~\ref{table:1}, our method achieves improvements of 25.3$\%$, 28.2$\%$, and 26.4$\%$ in the 1-shot task, respectively. Meanwhile, as shown in Table~\ref{table:2}, our method demonstrates improvements of 24.4$\%$, 29.4$\%$, and 25.0$\%$ in the 5-shot task, respectively. This result demonstrates the effectiveness of our proposed method, which significantly enhances the continual learning capabilities of existing Transformer-based policies. \\
On the other hand, we integrate classical continual learning methods (replay-based, regularization-based, prompt-based) into Transformer-based policies to enhance their continual learning capabilities. As shown in Table~\ref{table:1} and Table~\ref{table:2}, the replay-based, regularization-based, and prompt-based methods have improved the continual learning capabilities of existing Transformer-based policies. However, the performance of our proposed TPOIC remains notably superior to these approaches. Specifically, TPOIC demonstrates an average performance increase of 21.0$\%$ and 22.3$\%$ over the replay-based method in 1-shot and 5-shot tasks, respectively. Similarly, when compared to the regularization-based method, TPOIC exhibits an average performance gain of 16.4$\%$ and 15.6$\%$ in 1-shot and 5-shot tasks. Furthermore, compared to the prompt-based method, TPOIC also demonstrates an average performance improvement of 16.2$\%$ and 15.8$\%$ in 1-shot and 5-shot tasks, respectively. This result further illustrates that our proposed TPOIC is more suitable for few-shot action incremental learning tasks in embodied scenarios, demonstrating the superiority over other classical continual learning methods.\\
\textbf{Observations}.
These results lead us to three observations: \\
\textbf{a)} As the number of new tasks increases, the average accuracy of Transformer-based policies continues to decline, indicating severe catastrophic forgetting.\\
\textbf{b)} In some cases, under the 5-shot setting, although the model can better learn new tasks, it exhibits more pronounced forgetting of previous tasks, resulting in lower average accuracy compared to the 1-shot setting. \\
\textbf{c)} Notably, our proposed method not only preserves the skills learned in previous tasks but also effectively adapts to new tasks, achieving higher average accuracy. 
\subsection{Ablation Studies}
In this section, we thoroughly analyze the effectiveness of different components in TOPIC, i.e., TSP and CES, and we also examine the impact of different configurations on model performance.
\subsubsection{The Effectiveness of TSP and CES}
To investigate the impact of TSP and CES on model performance, we utilize existing Transformer-based policies as our baseline and analyze the importance of each component in the 1-shot task. As shown in Table~\ref{table:3}, leveraging only TSP without introducing the CES, the average model performance is improved by 14.6$\%$\
compared to baseline models. Furthermore, the introduction of the CES further enhances the model’s continual learning ability, leading to average improvements of 26.5$\%$ compared to baseline models. This result fully demonstrates that our proposed method can significantly enhance the continual learning ability of existing Transformer-based policies.
\begin{table}[t]
	\renewcommand{\arraystretch}{1.2}
	\setlength{\tabcolsep}{7.7pt}
	\centering
	\small
 \caption{The influence of varying combinations of coefficients $\lambda_1$ and $\lambda_2$. We report the average accuracy of 1-shot and 5-shot tasks of SAM-E and RVT across all sessions.}
	\scalebox{0.97}{
    
	\begin{tabular}{ccccc}
	\toprule
	\multirow{2}{*}{$\lambda$ Coefficient} & \multicolumn{2}{c}{SAM-E + TOPIC}  & \multicolumn{2}{c}{RVT + TOPIC}\\
	& 1-shot & 5-shot & 1-shot & 5-shot \\
	\midrule
	$\lambda_1$ = 0.1, $\lambda_2$ = 0.9 & 58.2 & 58.0  & \textbf{49.2} & 47.5 \\
	$\lambda_1$ = 0.2, $\lambda_2$ = 0.8 & \textbf{58.7} & \textbf{58.4} &  49.1  & \textbf{47.7}\\
	  $\lambda_1$ = 0.3, $\lambda_2$ = 0.7 & 55.6 & 56.5  & 47.1 & 44.5 \\
        $\lambda_1$ = 0.4, $\lambda_2$ = 0.6 & 50.1  &50.9  &  40.8 & 38.9 \\
        $\lambda_1$ = 0.5, $\lambda_2$ = 0.5 & 40.1 & 41.3  &  32.4 & 31.2 \\
	\bottomrule
\end{tabular}
}
	\label{table:5}
\end{table}
\begin{table}[t]
	\renewcommand{\arraystretch}{1.2}
	\setlength{\tabcolsep}{9pt}
	\centering
	\small
        \caption{Ablation experiment on different projection methods of average accuracy across all sessions. We report 1-shot and 5-shot tasks of SAM-E and RVT.}
	\scalebox{0.98}{
	\begin{tabular}{ccccc}

	\toprule
	\multirow{2}{*}{Projection method} & \multicolumn{2}{c}{SAM-E + TSP}  & \multicolumn{2}{c}{RVT + TSP}\\
	& 1-shot & 5-shot & 1-shot & 5-shot \\
	\midrule
	MLP & 44.8 & 45.0 & 36.3  & 37.5 \\
	Linear & 45.1  & \textbf{46.7} &  36.8   & 37.9\\
	 Average Pooling & \textbf{45.6} & 46.5  & \textbf{37.2} & \textbf{38.0}\\
	\bottomrule
\end{tabular}
}
	\label{table:6}
\end{table}
\begin{table}[t!]
\renewcommand{\arraystretch}{1.2}
	\setlength{\tabcolsep}{7.7pt}
	\small
	\centering
    \caption{Compared with the number of parameters and computation of the SAM-E baseline, the proposed TOPIC performs better.}
	\scalebox{0.98}{
		\begin{tabular}{cccccc}
			\toprule
			\multirow{2}{*}{Method} & \multirow{2}{*}{Params} & \multirow{2}{*}{GFLOPs}  & \multicolumn{2}{c}{FSAIL task}\\
			&  & & 1-shot & 5-shot\\
			\midrule
			SAM-E & 35.6M  & 412.52 & 30.5 & 29.0  \\
			TOPIC (Ours) & 35.5M & 412.52 & \textbf{58.7} & \textbf{58.4}  \\
			\bottomrule
		\end{tabular}
	}
    \label{table:7}
\end{table} 
\subsubsection{Number of Task-Specific Prompts}
\begin{figure}[t]
	\centering  
	\includegraphics[width=0.45\textwidth]{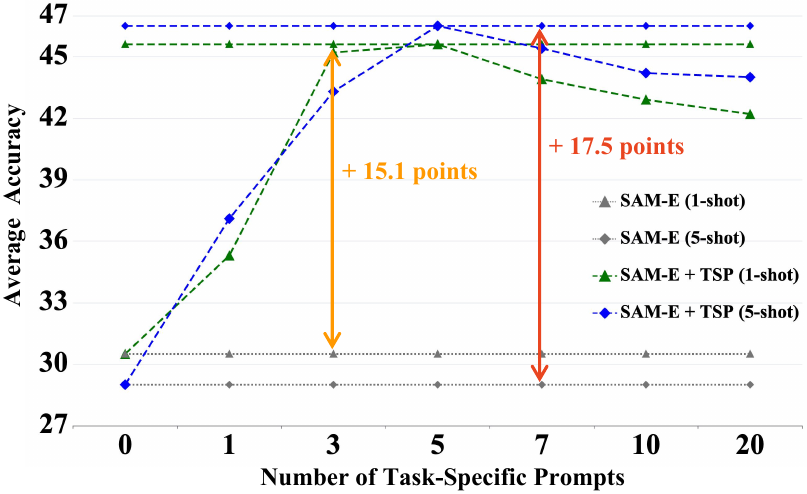}  
	\caption{Exploring the impact of the number of task-specific prompts. We report the average accuracy across all sessions.}
	\label{fig:5}  
\end{figure}

To investigate the influence of the number of predefined learnable task-specific prompts on model prediction accuracy, we leverage SAM-E as a baseline and examine the effects of different prompt quantities on model performance. As illustrated in Figure \ref{fig:5}, in both 1-shot and 5-shot tasks, the model achieves optimal performance when the number of prompts is set to five. Therefore, we default the number of task-specific prompts to five. Additionally, this result demonstrates that our proposed task-specific prompts significantly enhance the continual learning capabilities of existing Transformer-based policies. 
\subsubsection{Selection of the $\lambda$ Coefficient}
During the update process of our proposed Continuous Evolution Policy (CEP), we leverage coefficients $\lambda_1$ and $\lambda_2$ in Equation 3 to control the proportions of task-specific skills and generic skills, where a larger $\lambda_1$ indicates that the model retains more task-specific skills, and coefficient $\lambda_2$ is used to adjust the retention proportion of generic skills. As shown in Table~\ref{table:5}, we experiment with various combinations of these coefficients. TOPIC achieves the best results when  $\lambda_1$ is set to 0.2 and  $\lambda_2$ to 0.8. This demonstrates the importance of generic skills across different tasks, which further proves that the skills among tasks have intrinsic relationships and skills can be reused from previous tasks to new tasks.
\subsubsection{Structure of Task-Specific Prompt Projection.}
We attempt three different ways to implement Task-specific prompt projection: Average Pooling, Linear, and MLP. We conduct experiments on the 1-shot and 5-shot tasks on SAM-E~\cite{zhang2024sam} and RVT~\cite{goyal2023rvt}. As shown in Table~\ref{table:6}, SAM-E and RVT achieve optimal performance by employing the simplest method Average Pooling for processing the task-specific prompt tokens. This result demonstrates that the task-specific prompts interact deeply with multi-modal information from a few demonstrations, thereby capturing accurate and discriminative task-specific information.
\begin{figure}[t]
	\centering  
	\includegraphics[width=0.45\textwidth]{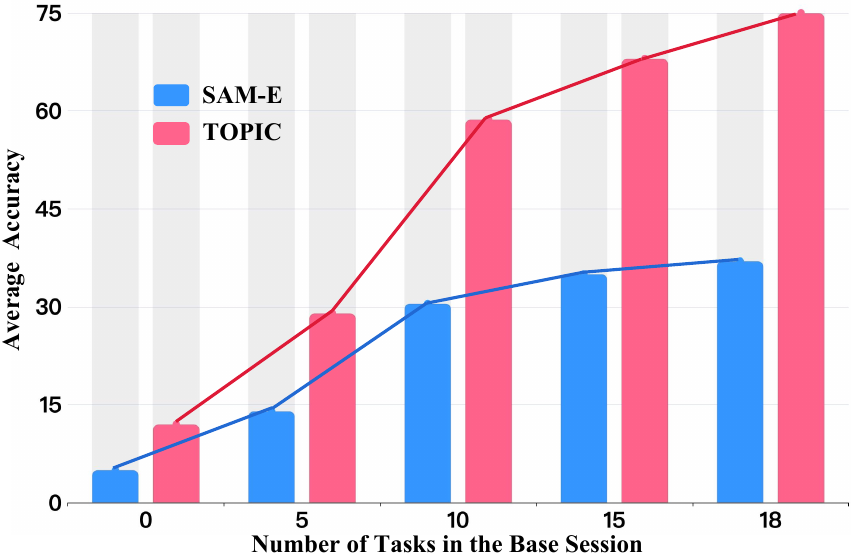}  
	\caption{Exploring the impact of the number of tasks in the base session. We report the average accuracy across all sessions.} 
	\label{fig:base_session}  
\end{figure} 
\subsubsection{The number of tasks in the base session}
Based on our theoretical analysis in Section III, as TOPIC learns various tasks, the more skills it can reuse, thereby enhancing the model's continual learning capability. To validate this hypothesis, we systematically investigate the impact of task diversity in the base learning session. We conduct comparative experiments on the 1-shot task against the baseline method SAM-E~\cite{zhang2024sam}. Specifically, as shown in Figure~\ref{fig:base_session}, the model's ability to continuously learn new tasks progressively improves with an increasing number of tasks in the base session. Notably, when the number of base session tasks is 0, TPOIC still performs better than the baseline model. This is because TOPIC possesses the ability to continuously learn new skills and reuse previously learned skills in the incremental session. This empirical evidence substantiates the scalability and adaptive learning potential of our proposed TPOIC, highlighting its robust skill transfer and generalization capabilities.
\subsubsection{Computation cost analysis}
We analyze the complexity of SAM-E and TOPIC. As shown in Table~\ref{table:7}, because our proposed TOPIC freezes the text encoder and visual encoder during the training process, it reduces the number of trainable parameters. In comparison, SAM-E employs LoRA~\cite{hu2021lora} to train the SAM~\cite{kirillov2023segment} encoder. Our proposed TOPIC has fewer trainable parameters and significantly outperforms SAM-E on the FSAIL task, which demonstrates that it is more suitable for continual learning embodied tasks with only a few demonstrations.
\subsection{More Detailed Experimental Analysis}
\begin{figure}[t]
        \flushleft
	\includegraphics[width=0.48\textwidth]{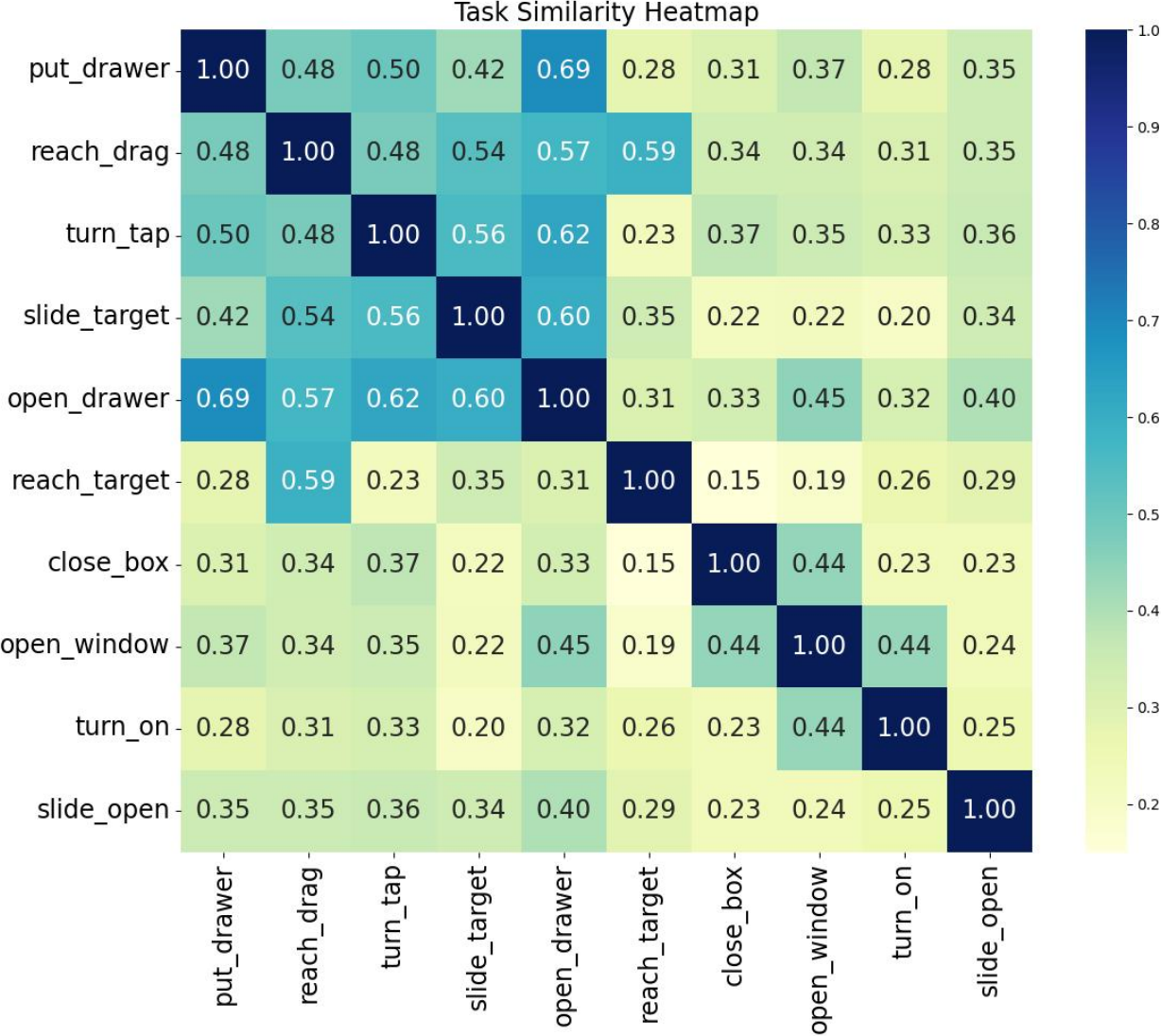}  
	\caption{Visualization of the intrinsic relationships of different task-specific prompts. }
	\label{fig:6}  
\end{figure} 
\subsubsection{Analysis of Task-Specific Prompts Relationships}
Based on the theoretical analyses in Sections III, our proposed task-specific prompts can extract task-specific discriminative information and construct a task relationship graph through the relationships between tasks, thereby facilitating skill reuse across different tasks. 
To empirically evaluate the task correlations, we visualize the intrinsic relationships among different tasks from the base session and incremental session. As shown in Figure~\ref{fig:6}, we calculate the cosine similarities between task-specific prompts learned for different tasks. In most cases, intuitively similar tasks exhibit higher similarity values of task-specific prompts. For instance, tasks such as ``open drawer" and ``put drawer" exhibit high similarity due to their shared  object (``drawer"), while tasks like ``reach target" and ``reach drag" are highly similar because of their shared action (``reach"). These results further validates that the task relationship graph constructed based on task-specific prompts can genuinely reflect the intrinsic connections between tasks. By enabling skill reuse across similar tasks, our proposed TPOIC allows Transformer-based policies to better adapt to new tasks by leveraging the skills learned from previous tasks.
\begin{table*}[p]
    \renewcommand\arraystretch{0.97}
    \begin{center}
        \setlength{\tabcolsep}{4pt}
        \centering
        \caption{FSAIL performance across tasks on \textbf{1-shot} setting. 
        ``Average Acc." is the average accuracy across all tasks. Mean of 5 evaluations are reported.}
        \label{table:10}
        \resizebox{0.97\textwidth}{!}{
            \begin{tabular}{lccccccccccccccccc}
                \toprule
                \multicolumn{1}{l}{\multirow{2}{*}{\bf Methods}}&\multicolumn{15}{c}{\bf Accuracy in each task (\%) $\uparrow$}&\bf Average&\bf Final \\ 
                \cmidrule{2-16}
                  &\bf 1&\bf 2&\bf 3&\bf 4&\bf 5&\bf 6&\bf 7&\bf 8&\bf 9&\bf 10&\bf 11&\bf 12&\bf 13&\bf 14&\bf 15&\bf Acc.&\bf Improv.  \\ 
                \midrule
                \multicolumn{18}{l}{\emph{Transformer-based policies: SAM-E~\cite{zhang2024sam}}} \\
                Session 0 &88.0&100.0&100.0&88.0&96.0&88.0&96.0&72.0&88.0&100.0&-&-&-&-&-&91.6& baseline\\
                Session 1 &0.0&0.0&100.0&25.0&52.0&0.0&56.0&0.0&0.0&0.0&56.0&-&-&-&-&26.3& baseline\\
                Session 2 &0.0&0.0&76.0&28.0&8.8&12.0&44.0&4.0&4.0&0.0&56.0&60.0&-&-&-&24.4& baseline\\
                Session 3 &0.0&0.0&64.0&16.0&0.0&8.0&32.0&0.0&0.0&0.0&44.0&28.0&44.0&-&-&18.2& baseline\\
                Session 4 &0.0&0.0&28.0&0.0&4.0&4.0&4.0&0.0&4.0&0.0&32.0&4.0&20.0&60.0&-&11.4& baseline\\
                Session 5 &0.0&8.0&40.0&4.0&0.0&0.0&8.0&0.0&0.0&0.0&52.0&0.0&0.0&12.0&40.0&10.9& baseline\\
                \midrule
                \multicolumn{18}{l}{\emph{Transformer-based policies + Regularization~\cite{kirkpatrick2017overcoming}}} \\
                Session 0 &88.0&100.0&100.0&88.0&96.0&88.0&96.0&72.0&88.0&100.0&-&-&-&-&-&91.6& + 0.0\\
                Session 1 &40.0 & 20.0 & 80.0 & 44.0 & 56.0 & 48.0 & 33.6 & 0.0 & 20.8 & 0.0 & 56.0&-&-&-&-&36.2&  + 9.9\\
                Session 2 & 0.0 & 16.0 & 80.0 & 52.8 & 44.0 & 28.0 & 52.0 & 12.0 & 22.4 & 0.0 & 56.0 & 48.0&-&-&-&34.3& + 9.9 \\
                Session 3 &0.0 & 20.0 & 56.0 & 36.0 & 52.0 & 20.0 & 56.0 & 0.0 & 64.0 & 12.8 & 44.0 & 56.0 & 8.0&-&-&32.7& + 14.5\\
                Session 4 &0.0 & 16.0 & 24.0 & 20.0 & 36.0 & 32.0 & 52.0 & 0.0 & 36.0 & 8.0 & 64.0 & 44.0 & 48.0 & 32.0&-&29.4& + 18.0\\
                Session 5 &0.0 & 12.0 & 20.0 & 32.8 & 20.0 & 20.0 & 32.0 & 0.0 & 32.0 & 4.0 & 52.0 & 56.0 & 32.0 & 64.0 & 20.0&26.5& + 15.6\\
                \midrule
                \multicolumn{18}{l}{\emph{\textbf{Ours: Transformer-based policies + TSP } }}\\
                Session 0 &84.0&100.0&100.0&96.0&96.0&91.2&100.0&68.0&84.0&100.0&-&-&-&-&-&91.9& + 0.3\\
                Session 1 &0.0 & 0.0 & 96.0 & 59.2 & 63.2 & 84.0 & 100.0 & 80.0 & 92.0 & 0.0 & 48.0&-&-&-&-&56.6& + 30.3\\
                Session 2 &0.0 & 0.0 & 92.0 & 56.0 & 88.0 & 27.2 & 92.0 & 44.0 & 40.0 & 4.0 & 52.0 & 64.0&-&-&-&46.6& + 22.2\\
                Session 3 &12.0 & 0.0 & 88.0 & 50.4 & 80.0 & 0.0 & 92.0 & 40.0 & 4.0 & 20.0 & 52.0 & 8.0 & 40.0&-&-&37.4& + 19.2\\
                Session 4 &4.0 & 0.0 & 80.0 & 36.0 & 60.0 & 0.0 & 68.0 & 12.0 & 12.0 & 0.0 & 52.0 & 0.0 & 0.0 & 12.0&-&24.0& + 12.6\\
                Session 5 &0.0 & 0.0 & 48.0 & 8.0 & 0.0 & 4.0 & 20.0 & 0.0 & 4.0 & 0.0 & 24.0 & 8.0 & 32.0 & 28.0 & 76.0&16.8& + 5.9\\
                \midrule
                \multicolumn{18}{l}{\emph{\textbf{Ours: Transformer-based policies + TOPIC} }}\\
                Session 0 &84.0&100.0&100.0&96.0&96.0&91.2&100.0&68.0&84.0&100.0&-&-&-&-&-&\textbf{91.9}& \textbf{+ 0.3}\\
                Session 1 &92.0&48.0&100.0&76.8&96.0&92.0&100.0&52.0&80.0&28.0&48.0&-&-&-&-&\textbf{73.9}& \textbf{+ 47.6}\\
                Session 2 &56.0&28.0&96.0&80.0&81.6&88.0&100.0&68.0&76.0&0.0&52.0&52.0&-&-&-&\textbf{64.8}& \textbf{+ 40.4}\\
                Session 3 &17.6&56.0&88.0&72.0&84.0&76.0&84.0&55.2&32.0&4.0&60.0&8.0&4.0&-&-&\textbf{49.3}& \textbf{+ 31.1}\\
                Session 4 &0.0&88.0&83.2&40.0&44.0&16.0&80.0&74.4&48.0&8.0&40.0&0.0&0.0&32.0&-&\textbf{39.5}& \textbf{+ 28.1}\\
                Session 5 &0.0&68.0&80.0&35.2&52.0&0.0&92.0&60.0&12.0&0.0&36.0&0.0&8.0&20.0&28.0&\textbf{32.7}& \textbf{+ 21.8}\\
                \bottomrule
            \end{tabular}
        }
    \end{center}

    \vfill

    \renewcommand\arraystretch{1.0}
    \begin{center}
        \setlength{\tabcolsep}{4pt}
        \centering
        \caption{FSAIL performance across tasks on \textbf{5-shot} setting. 
        ``Average Acc." is the average accuracy across all tasks. Mean of 5 evaluations are reported.}
        \label{table:11}
        \resizebox{0.97\textwidth}{!}{
            \begin{tabular}{lccccccccccccccccc}
                \toprule
                \multicolumn{1}{l}{\multirow{2}{*}{\bf Methods}}&\multicolumn{15}{c}{\bf Accuracy in each task (\%) $\uparrow$}&\bf Average&\bf Final \\ 
                \cmidrule{2-16}
                  &\bf 1&\bf 2&\bf 3&\bf 4&\bf 5&\bf 6&\bf 7&\bf 8&\bf 9&\bf 10&\bf 11&\bf 12&\bf 13&\bf 14&\bf 15&\bf Acc.&\bf Improv.  \\ 
                \midrule
                \multicolumn{18}{l}{\emph{Transformer-based policies: SAM-E~\cite{zhang2024sam}}} \\
                Session 0 &88.0&100.0&100.0&88.0&96.0&88.0&96.0&72.0&88.0&100.0&-&-&-&-&-&91.6& baseline\\
                Session 1 &0.0&0.0&100.0&32.0&28.0&0.0&50.4&0.0&0.0&0.0&68.0&-&-&-&-&25.3& baseline\\
                Session 2 &0.0&0.0&76.0&16.0&0.0&0.0&20.0&0.0&0.0&0.0&64.0&76.0&-&-&-&21.0& baseline\\
                Session 3 &0.0&0.0&79.2&0.0&0.0&4.0&12.0&0.0&0.0&0.0&44.0&12.0&32.0&-&-&14.1& baseline\\
                Session 4 &0.0&0.0&28.0&0.0&4.0&0.0&4.0&0.0&4.0&0.0&32.0&8.0&20.0&64.0&-&11.7& baseline\\
                Session 5 &0.0&12.0&36.0&4.0&0.0&0.0&32.0&0.0&0.0&0.0&16.0&0.0&0.0&8.0&44.0&10.1& baseline\\
                \midrule
                \multicolumn{18}{l}{\emph{Transformer-based policies  + Regularization~\cite{kirkpatrick2017overcoming}}} \\
                Session 0 &88.0&100.0&100.0&88.0&96.0&88.0&96.0&72.0&88.0&100.0&-&-&-&-&-&91.6& + 0.0\\
                Session 1 &0.0 & 16.0 & 88.0 & 44.0 & 24.0 & 0.0 & 52.0 & 36.0 & 48.0 & 0.0 & 68.0&-&-&-&-&34.2& + 8.9\\
                Session 2 &0.0 & 12.0 & 52.0 & 48.0 & 26.8 & 8.8 & 52.0 & 20.0 & 48.0 & 4.0 & 72.0 & 56.0&-&-&-&33.3& + 12.3\\
                Session 3 &0.0 & 0.0 & 52.0 & 32.0 & 24.0 & 4.0 & 56.0 & 12.0 & 40.0 & 4.0 & 51.2 & 64.0 & 16.0&-&-&30.1& + 16.0\\
                Session 4 &0.0 & 8.0 & 40.0 & 32.0 & 24.0 & 16.0 & 52.0 & 4.0 & 32.0 & 8.0 & 48.0 & 56.0 & 64.0 & 24.0&-&29.4& + 17.7\\
                Session 5 &0.0 & 0.0 & 36.0 & 24.8 & 20.0 & 8.0 & 44.0 & 8.0 & 40.0 & 4.0 & 32.0 & 52.0 & 52.0 & 68.0 & 32.0&28.1& + 18.0\\
                \midrule
                \multicolumn{18}{l}{\emph{\textbf{Ours: Transformer-based policies + TSP } }}\\
                Session 0 &84.0&100.0&100.0&96.0&96.0&91.2&100.0&68.0&84.0&100.0&-&-&-&-&-&91.9& + 0.3\\
                Session 1 &0.0&0.0&88.0&52.0&36.0&88.0&100.0&68.0&76.0&0.0&56.0&-&-&-&-&51.3& + 26.0\\
                Session 2 &0.0&0.0&88.0&28.0&92.0&4.0&100.0&32.0&16.0&0.0&76.0&72.0&-&-&-&42.3& + 21.3\\
                Session 3 &12.0&0.0&92.0&36.0&88.0&4.0&92.0&40.0&20.0&0.0&60.0&20.0&16.0&-&-&36.9& + 22.8\\
                Session 4 &0.0&0.0&68.0&56.0&72.0&0.0&72.0&12.0&12.0&0.0&64.0&0.0&0.0&72.0&-&30.6&+ 18.9\\
                Session 5 &0.0&0.0&80.0&16.0&8.0&0.0&92.0&8.0&4.0&0.0&40.0&12.0&4.0&52.0&76.0&26.1& + 15.0\\
                \midrule
                \multicolumn{18}{l}{\emph{\textbf{Ours: Transformer-based policies + TOPIC }}}\\
                Session 0 &84.0&100.0&100.0&96.0&96.0&91.2&100.0&68.0&84.0&100.0&-&-&-&-&-&\textbf{91.9}& \textbf{+ 0.3}\\
                Session 1 &68.0&28.0&88.0&72.0&92.0&84.0&100.0&84.0&68.0&4.0&56.0&-&-&-&-&\textbf{67.6}& \textbf{+ 42.3}\\
                Session 2 &56.0&28.0&96.0&72.0&84.0&84.8&100.0&68.0&76.0&0.0&48.0&56.0&-&-&-&\textbf{64.1}& \textbf{+ 43.1}\\
                Session 3 &36.0&60.0&100.0&63.2&96.0&84.0&96.0&40.0&32.0&0.0&76.0&12.0&24.0&-&-&\textbf{55.3}& \textbf{+ 41.2}\\
                Session 4 &20.0&52.0&56.0&40.0&44.0&32.0&88.0&76.0&56.0&28.0&80.0&4.0&0.0&28.0&-&\textbf{43.1}& \textbf{+ 31.4}\\
                Session 5 &0.0&44.0&68.0&15.2&8.0&44.0&60.0&40.0&15.2&4.0&44.0&4.0&0.0&44.0&36.0&\textbf{28.4}& \textbf{+ 18.3}\\
                \bottomrule
            \end{tabular}
        }
    \end{center}
\end{table*}
\begin{figure}[t]
	\centering  
	\includegraphics[width=0.38\textwidth]{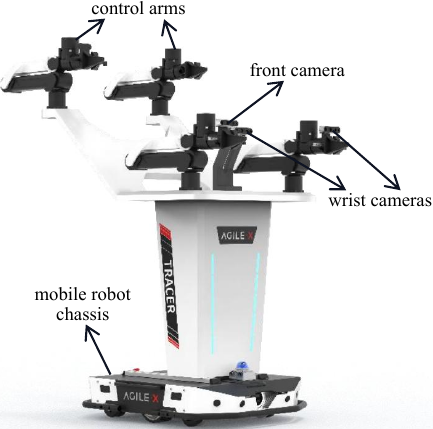}  
	\caption{Real-World hardware configuration. We use the Cobot Mobile ALOHA, which is equipped with two control arms, two slave arms, two wrist cameras and a frontal camera.}  
	\label{fig:7}  
\end{figure} 
\begin{table}[t]
    \renewcommand{\arraystretch}{1.35}
    \setlength{\tabcolsep}{3pt}
    \centering
    \caption{Results in real-world FSAIL tasks. We report the 1-shot task of SAM-E and TOPIC. }
    \scalebox{0.9}{
    \begin{tabular}{lcccccccl}
        \toprule
        \multicolumn{1}{l}{\multirow{2}{*}{\bf Methods}} & \multicolumn{6}{c}{\bf Accuracy in each session (\%) $\uparrow$} &\bf Average &\bf Final\\ 
        \cmidrule(lr){2-7}
        & \bf 0 & \bf 1 & \bf 2 & \bf 3 & \bf 4 &\bf 5 & \bf Acc. &\bf Improv.\\
        \midrule
        \multicolumn{8}{l}{\emph{Transformer-based policies}} \\
        SAM-E~\cite{zhang2024sam} & 47.0   &15.5 &12.5 &10.0 &6.4 & 3.3            &15.8      &baseline   \\
        \midrule
        \multicolumn{8}{l}{\emph{\textbf{Ours: Transformer-based policies  + \textbf{TOPIC}} }} \\

SAM-E~\cite{zhang2024sam} + \textbf{TOPIC} & \textbf{49.0} & \textbf{32.7} & \textbf{24.2} & \textbf{18.5} & \textbf{13.4} & \textbf{10.0} & \textbf{24.6} & \textbf{+ 8.8 } \\

        \bottomrule
    \end{tabular}
}
    \label{table:4}
\end{table}
\subsubsection{Performance of different tasks}
To provide a more detailed experimental comparison, we compare the performance of each task with the baseline SAM-E~\cite{zhang2024sam} and the regularization-based method~\cite{kirkpatrick2017overcoming}, including 10 tasks from the base session and 5 tasks from the incremental sessions. As shown in Table~\ref{table:10} and Table~\ref{table:11}, in each incremental session, SAM-E and the regularization-based method only learn the skills of new tasks while exhibiting significant catastrophic forgetting of the previous learned skills. However, our proposed TOPIC not only adapts to new tasks but also retains the skills acquired from previous tasks, achieving a higher average accuracy across all tasks compared to other methods and effectively mitigating the issue of catastrophic forgetting.
\subsection{Real-World Experiments}
\begin{figure*}[t] 
	\centering 
	\includegraphics[width=1.0\textwidth]{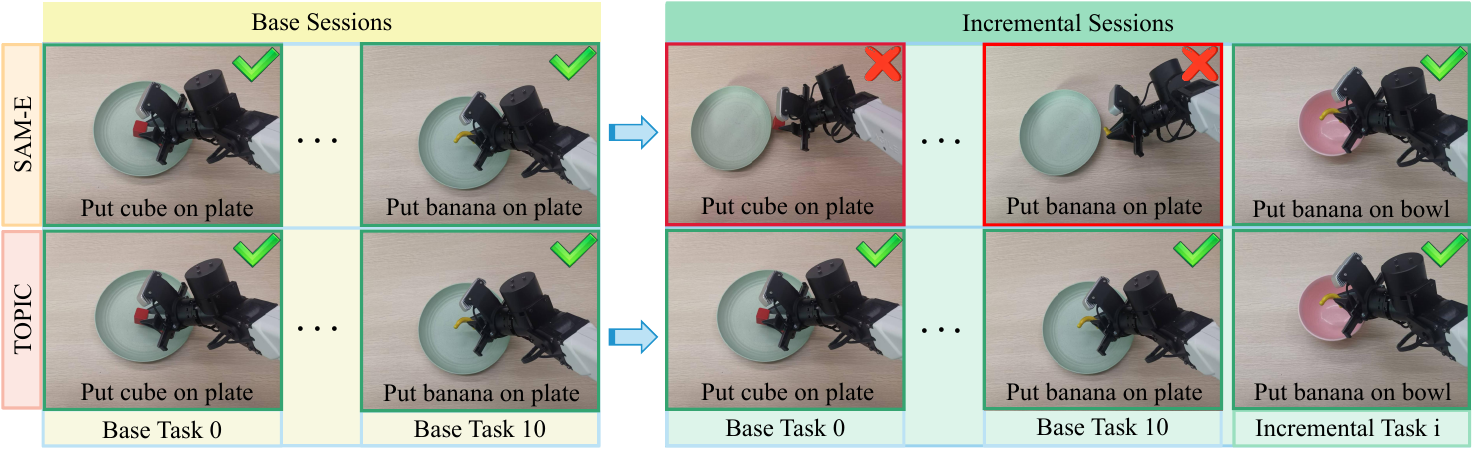} 
    	\caption{Qualitative Results in Real-world. We design 10 tasks in the base session and 5 tasks in the incremental session to validate the model's continual learning capability with novel objects or actions. TOPIC can avoid catastrophic forgetting of previous learned tasks while adapting to new tasks. } 
	\label{fig:4} 
\end{figure*}
\subsubsection{Task Setting in Real-World}
Similarly to the simulated environment, we also set up FSAIL tasks in real world scenarios. Specifically, we first design and collect ten different real-world tasks and each task with 50 demonstrations to serve as our base session. Then, we collect five unseen tasks and each task with only five demonstrations to form the incremental session, which tests the model's continual learning capability with few-shot demonstrations. 
Specifically, in order to more comprehensively explore the model's continuous learning ability, we define tasks involving different actions and objects, and tasks in incremental sessions being unseen in the base session. For each task, we perform ten times inference and calculate the average accuracy to validate the continual learning capability of different methods.
\subsubsection{Hardware Information}
We provide a description of the hardware configuration of our real-world experiments. We deploy the SAM-E~\cite{zhang2024sam} baseline and our proposed TOPIC on the Cobot Mobile ALOHA, a robot utilizing the Mobile ALOHA system architecture~\cite{fu2024mobile}. The robot’s appearance is shown in Figure \ref{fig:7}, and it is equipped with two control arms, two slave arms, two wrist cameras and a frontal camera. It is crucial to emphasize that we only use a single 6 DoF robotic arm during our experiments, and all tasks are static, without leveraging any of the robot’s mobility capabilities. 
\subsubsection{Comparison with other methods}
To demonstrate the effectiveness of our proposed TPOIC in real-world scenarios, we train and test TOPIC employing the Cobot Mobile ALOHA~\cite{zhao2023learning}. We compare our proposed TOPIC with the SAM-E baseline, where each task is tested ten times, and the average accuracy is reported. Figure \ref{fig:4} and Table~\ref{table:4} show qualitative and quantitative comparisons in real-world scenarios, respectively. These results demonstrates that our proposed TOPIC notably improves the continual learning capability of baseline model.
\section{Conclusion}
Our work pioneers few-shot continual learning in robotic manipulation tasks, introducing the \textbf{F}ew-\textbf{S}hot \textbf{A}ction-\textbf{I}ncremental \textbf{L}earning (\textbf{FSAIL}) setting and accordingly designing a \textbf{T}ask-pr\textbf{O}mpt gra\textbf{P}h evolut\textbf{I}on poli\textbf{C}y (\textbf{TOPIC}). Specifically, we introduce \textbf{T}ask-\textbf{S}pecific \textbf{P}rompts (\textbf{TSP}), which extracts task-specific information from a few demonstrations to guide action prediction, effectively mitigating the issue of data scarcity in embodied tasks. Meanwhile, we propose a \textbf{C}ontinuous \textbf{E}volution \textbf{S}trategy (\textbf{CES}) that leverages the intrinsic connections between tasks to construct a task relation graph. This graph enables the reuse of learned skills during the adaptation to new tasks, thereby mitigating the problem of catastrophic forgetting. TOPIC significantly enhances the continual learning capabilities of existing Transformer-based policies with few-shot demonstrations. Comparison experiments and ablation studies demonstrate the effectiveness of our proposed method. We hope that our innovations can inspire and promote the FSAIL research.
\section{Limitation and Future Work}
In this paper, we propose the TOPIC to adapt to new tasks with few-shot demonstrations while mitigating the issue of catastrophic forgetting. TOPIC can be seamlessly integrated with existing Transformer-based policies, significantly enhancing their continual learning capabilities. However, we also identify several limitations that highlight potential directions for future research. First, due to computational resource constraints, we are unable to expand the diversity of the base tasks or increase the scale of the model. We believe that increasing the number of base tasks and scaling up the model size can enhance adaptability to new tasks. Second, we observed a substantial gap between simulation and real-world performance. Future work will focus on addressing continual learning challenges within real-world contexts. Third, although our method enhances the continual learning performance of existing Transformer-based policies under the constraint of limited demonstrations, there is still a long way to go before we can effectively adapt to new tasks with few-shot demonstrations and fully address the issue of catastrophic forgetting in previously learned tasks. Future investigations should prioritize addressing the nuanced challenges of few-shot action-incremental learning, focusing on developing more sophisticated strategies that can effectively balance skill retention and adaptive learning. We anticipate that continued exploration in this domain will yield critical insights into overcoming the fundamental limitations of current continual learning paradigms in embodied AI.
\newpage

\bibliographystyle{IEEEtran}
\bibliography{main}


 





\end{document}